%% file: arxiv.tex
\documentclass[10pt,twocolumn,letterpaper]{article}

\usepackage{iccv}
\usepackage{times}
\usepackage{epsfig}
\usepackage{graphicx}
\usepackage{amsmath}
\usepackage{amssymb}
\usepackage{booktabs}
\usepackage{multirow}

\usepackage{subcaption}

\newcommand*{\affaddr}[1]{#1} 
\newcommand*{\affmark}[1][*]{\textsuperscript{#1}}

\usepackage[breaklinks=true,bookmarks=false]{hyperref}

 \iccvfinalcopy 

\def\iccvPaperID{1971} 

\ificcvfinal\fi
\begin{document}

\title{Semi-supervised Domain Adaptation via Minimax Entropy}
\author{%
  Kuniaki Saito\affmark[1], Donghyun Kim\affmark[1], Stan Sclaroff\affmark[1], Trevor Darrell\affmark[2] and Kate Saenko\affmark[1]\\
  \affaddr{\affmark[1]Boston University}, \affaddr{\affmark[2]University of California, Berkeley}\\
  \tt\small $^1$\{keisaito, donhk, sclaroff, saenko\}@bu.edu, $^2$trevor@eecs.berkeley.edu}
\maketitle
\ificcvfinal\thispagestyle{empty}\fi

\begin{abstract}

Contemporary domain adaptation methods are very effective at aligning feature distributions of source and target domains without any target  supervision. However, we show that these techniques perform poorly when even a few labeled examples are available in the target domain. To address this semi-supervised domain adaptation (SSDA) setting, we propose a novel Minimax Entropy (MME) approach that adversarially optimizes an adaptive few-shot model. Our base model consists of a feature encoding network, followed by a classification layer that computes the features' similarity to estimated prototypes (representatives of each class). Adaptation is achieved by alternately maximizing the conditional entropy of unlabeled target data with respect to the classifier and minimizing it with respect to the feature encoder. We empirically demonstrate the superiority of our method over many baselines, including conventional feature alignment and few-shot methods, setting a new state of the art for SSDA. Our code is available at \url{http://cs-people.bu.edu/keisaito/research/MME.html}.
 
\end{abstract}
\input{introduction.tex}
\input{related.tex}
\input{method.tex}

\input{experiment.tex}

\input{conclusion.tex}

{\small
\bibliographystyle{ieee_fullname}
\bibliography{egbib}
}
\input{supp_arxiv.tex}

\end{document}

%% file: introduction.tex
\section{Introduction}
\begin{figure}[t]
\begin{center}
   \includegraphics[width=\linewidth]{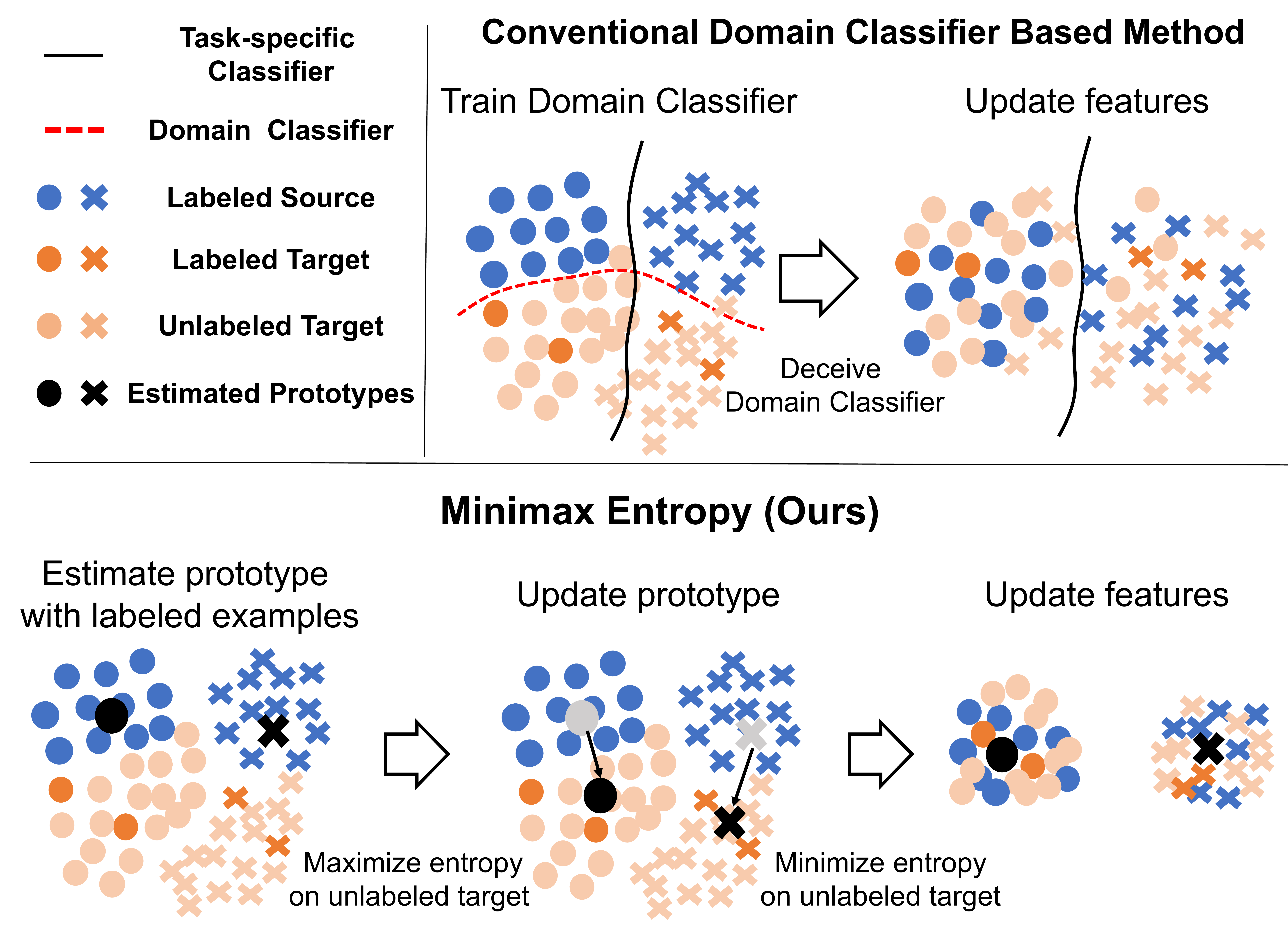}
\end{center}
\vspace{-6mm}
   \caption{\small We address the task of semi-supervised domain adaptation. 
   Top: Existing domain-classifier based methods align source and target distributions but can fail by
   generating ambiguous features near the task decision boundary. Bottom: Our method estimates a representative point of each class (prototype) and extracts discriminative features using a novel minimax entropy technique.}
\label{fig:fig1}
\end{figure}

Deep convolutional neural networks~\cite{krizhevsky2012imagenet} have significantly improved image classification accuracy with the help of large quantities of labeled training data, but  often generalize poorly to new domains. Recent unsupervised domain adaptation (UDA) methods~\cite{ganin2014unsupervised, long2015learning,long2017conditional,ADR,tzeng2014deep} improve generalization on unlabeled target data by aligning distributions, but can fail to learn discriminative class boundaries on target domains (see Fig.~\ref{fig:fig1}.) We show that in the Semi-Supervised Domain Adaptation (SSDA) setting where a few target labels are available, such methods often do not improve performance relative to just training on labeled source and target examples, and can even make it worse.
\begin{figure*}[t]
\begin{center}
   \includegraphics[width=\linewidth]{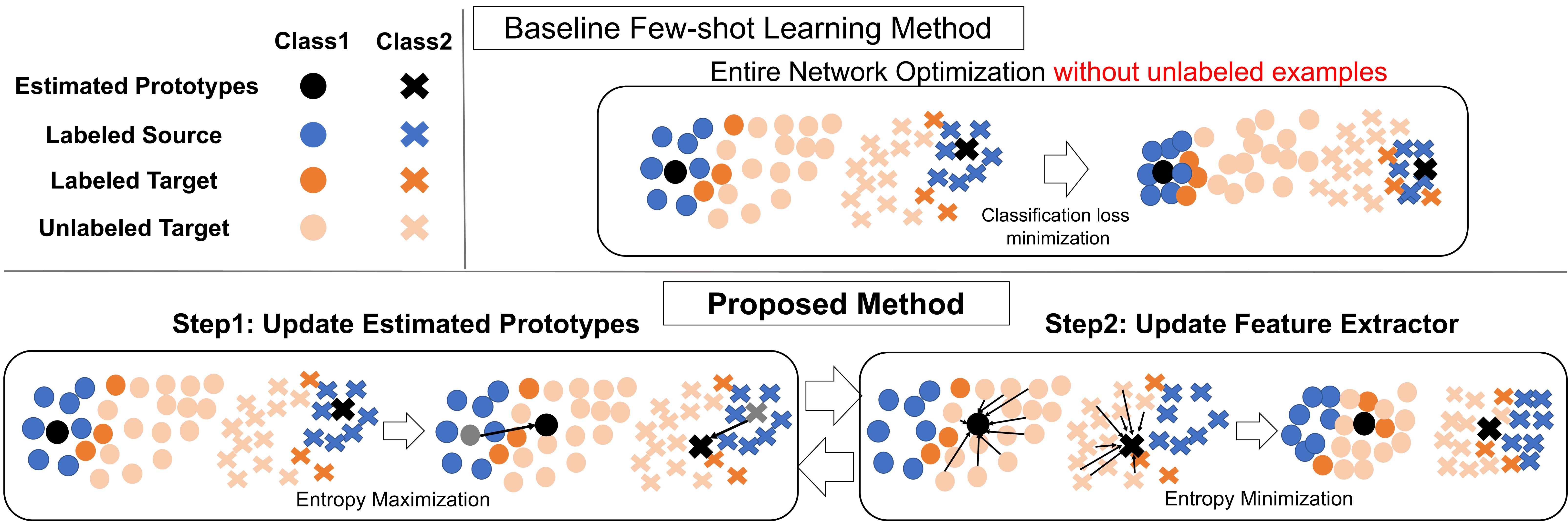}
\end{center}
\vspace{-6mm}
   \caption{
   Top: baseline few-shot learning method, which estimates class prototypes by weight vectors, yet does not consider unlabeled data. Bottom: our model extracts discriminative and domain-invariant features using unlabeled data through a domain-invariant prototype estimation. Step 1: we update the estimated prototypes in the classifier to maximize the entropy on the unlabeled target domain. Step 2: we minimize the entropy with respect to the feature extractor to cluster features around the estimated prototype.}
\label{fig:entropy-minmax}
\end{figure*}

We propose a novel approach for SSDA that overcomes the limitations of previous methods and significantly improves the accuracy of deep classifiers on novel domains with only a few labels per class. 
Our approach, which we call Minimax Entropy (MME), is based on optimizing a minimax loss on the conditional entropy of unlabeled data, as well as the task loss; this reduces the distribution gap while learning discriminative features for the task. 


We exploit a cosine similarity-based classifier architecture
recently proposed for few-shot learning~\cite{gidaris2018dynamic,chen2018closer}. The classifier (top layer) predicts a K-way class probability vector by computing
 cosine similarity between K class-specific weight vectors and the output of a feature extractor (lower layers), followed by a softmax.
Each class weight vector is an estimated ``prototype'' that can be regarded as a representative point of that class.
While this approach outperformed more advanced methods in few-shot learning and we confirmed its effectiveness in our setting, 
as we show below it is still quite limited. In particular, it does not leverage unlabeled data in the target domain. 

Our key idea is to minimize the distance between the class prototypes and neighboring unlabeled target samples, thereby extracting discriminative features. The problem is how to estimate domain-invariant prototypes without many labeled target examples.
The prototypes are dominated by the source domain, as shown in the leftmost side of Fig.~\ref{fig:entropy-minmax} (bottom), as the vast majority of labeled examples come from the source. 
To estimate domain-invariant prototypes, we move weight vectors toward the target feature distribution.
Entropy on target examples represents the similarity between the estimated prototypes and target features. 
A uniform output distribution with high entropy indicates that the examples are similar to all prototype weight vectors. Therefore, we move the weight vectors towards target by maximizing the entropy of unlabeled target examples in the first adversarial step. Second, we update the feature extractor to minimize the entropy of the unlabeled examples, to make them better clustered around the prototypes. This process is formulated as a mini-max game between the weight vectors and the feature extractor and applied over the unlabeled target examples.

Our method offers a new state-of-the-art in performance on SSDA; as reported below, we reduce the error relative to baseline few-shot methods which ignore unlabeled data by 8.5\%, relative to current best-performing alignment methods by 8.8\%, and relative to a simple model jointly trained on source and target by 11.3\% in one adaptation scenario.
Our contributions are summarized as follows:
\begin{itemize}
\item We highlight the limitations of  state-of-the-art domain adaptation methods 
\vspace{-3mm}
in the SSDA setting;
\item We propose a novel adversarial method, Minimax Entropy (MME),  designed for the SSDA task;
\vspace{-3mm}
\item We show our method's superiority to existing methods on benchmark datasets for domain adaptation.
\end{itemize}

%% file: related.tex
\section{Related Work}
\vspace{-3mm}
\textbf{Domain Adaptation.}
Semi-supervised domain adaptation (SSDA) is a very important task~\cite{donahue2013semi,yao2015semi,ao2017fast}, however it has not been fully explored, especially with regard to deep learning based methods. We revisit this task and compare our approach to recent semi-supervised learning or unsupervised domain adaptation methods. 
The main challenge in domain adaptation (DA) is the gap in feature distributions between domains, which degrades the source classifier's performance.
Most recent work has focused on unsupervised domain adaptation (UDA) and, in particular, feature distribution alignment. The basic approach measures the distance between feature distributions in source and target, then trains a model to minimize this distance. Many UDA methods utilize a domain classifier to measure the distance~\cite{ganin2014unsupervised,tzeng2014deep,long2015learning,long2017conditional,shu2018dirt}. The domain-classifier is trained to discriminate whether input features come from the source or target, whereas the feature extractor is trained to deceive the domain classifier to match feature distributions. 
UDA has been applied to various applications such as image classification~\cite{saenko2010}, semantic segmentation~\cite{learnfromsynth}, and object detection~\cite{dafaster,saito2018strong}. 
Some methods minimize task-specific decision boundaries' disagreement on target examples~\cite{saito2017maximum,ADR} to push target features far from decision boundaries. In this respect, they increase between-class variance of target features; on the other hand, we propose to make target features well-clustered around estimated prototypes. Our MME approach can reduce within-class variance as well as increasing between-class variance, which results in more discriminative features. Interestingly, we empirically observe that UDA methods~\cite{ganin2014unsupervised,long2017conditional,ADR} often fail in improving accuracy in SSDA. 


\begin{figure*}[t!]
\begin{center}
   \includegraphics[width=0.95\linewidth]{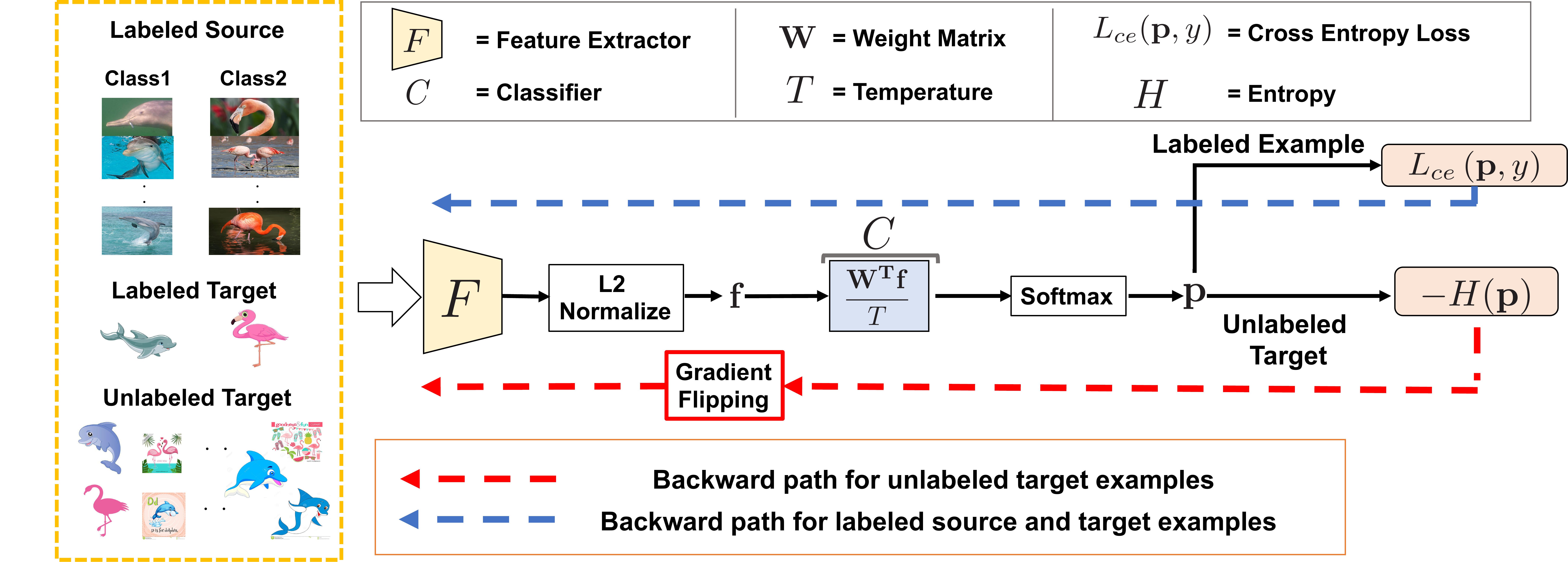}
\end{center}
\vspace{-4mm}
   \caption{An overview of the model architecture and MME. The inputs to the network are labeled source examples ($y$=label), a few labeled target examples, and unlabeled target examples. Our model consists of the feature extractor $F$ and the classifier $C$ which has weight vectors ($\mathbf{W}$) and temperature $T$. $\mathbf{W}$ is trained to maximize entropy on unlabeled target (Step 1 in Fig. ~\ref{fig:entropy-minmax}) whereas $F$ is trained to minimize it (Step 2 in Fig. ~\ref{fig:entropy-minmax}). To achieve the adversarial learning, the sign of gradients for entropy loss on unlabeled target examples is flipped by a gradient reversal layer~\cite{ganin2014unsupervised,tzeng2014deep}.}
\label{fig:pipeline}
\end{figure*}

\textbf{Semi-supervised learning (SSL).}
Generative~\cite{dai2017good,salimans2016improved}, model-ensemble~\cite{laine2016temporal}, and adversarial approaches~\cite{miyato2015distributional} have boosted performance in semi-supervised learning, but do not address domain shift. 
Conditional entropy minimization (CEM) is a widely used method in SSL~\cite{grandvalet2005semi, erkan2010semi}. However, we found that CEM fails to improve performance when there is a large domain gap between the source and target domains (see experimental section.)
MME can be regarded as a variant of entropy minimization which overcomes the limitation of CEM in domain adaptation.

\textbf{Few-shot learning (FSL).} 
Few shot learning ~\cite{snell2017prototypical,vinyals2016matching,ravi2016optimization} aims to learn novel classes given a few labeled examples and labeled ``base'' classes. SSDA and FSL make different assumptions: FSL does not use unlabeled examples and aims to acquire knowledge of \textit{novel} classes, while SSDA aims to adapt to the \textit{same} classes in a new domain. However both tasks aim to extract discriminative features given a few labeled examples from a novel domain or novel classes. We employ a network with $\ell_2$ normalization on features before the last linear layer and a temperature parameter $T$, which was proposed for face verification~\cite{ranjan2017l2} and applied to few-shot learning~\cite{gidaris2018dynamic, chen2018closer}. Generally, classification of a feature vector with a large norm results in confident output. To make the output more confident, networks can try to increase the norm of features. However, this does not necessarily increase the between-class variance because increasing the norm does not change the direction of vectors. $\ell_2$ normalized feature vectors can solve this issue. To make the output more confident, the network focuses on making the direction of the features from the same class closer to each other and separating different classes. This simple architecture was shown to be very effective for few-shot learning~\cite{chen2018closer} and we build our method on it in our work. 

%% file: method.tex
\section{Minimax Entropy Domain Adaptation}
\vspace{-1mm}
In semi-supervised domain adaptation, we are given source images and the corresponding labels in the source domain $\mathcal{D}_{s}=\left\{\left(\mathbf{x}_{i}^{s}, {y_{i}}^{s}\right)\right\}_{i=1}^{m_{s}}$.  In the target domain, we are also given a limited number of labeled target images $\mathcal{D}_{t}=\left\{\left(\mathbf{x}_{i}^{t}, {y_{i}}^{t}\right)\right\}_{i=1}^{m_{t}}$, as well as unlabeled target images $\mathcal{D}_{u} = \left\{\left(\mathbf{x}_{i}^{u} \right)\right\}_{i=1}^{m_{u}}$. Our goal is to train the model on $\mathcal{D}_{s}, \mathcal{D}_{t}, \text{and } \mathcal{D}_{u}$ and evaluate on $ \mathcal{D}_{u}$.
\subsection{Similarity based Network Architecture}
\vspace{-1mm}
Inspired by~\cite{chen2018closer}, our base model consists of a feature extractor $F$ and a classifier $C$. For the feature extractor $F$, we employ a deep convolutional neural network and perform $\ell_2$ normalization on the output of the network. 
Then, the normalized feature vector is used as an input to $C$ which consists of weight vectors $\mathbf{W}=[\mathbf{w}_1, \mathbf{w}_2, \dots, \mathbf{w}_K]$ where $K$ represents the number of classes and a temperature parameter $T$. $C$ takes $\frac{F(\mathbf{x})}{\|F(\mathbf{x})\|}$ as an input and outputs $\small{\frac{1}{T}\frac{\mathbf{W}^{\mathrm{T}}F(\mathbf{x})}{\|F(\mathbf{x})\|}}$. 
The output of C is fed into a softmax-layer to obtain the probabilistic output $\mathbf{p}\in R^{n}$. We denote $\mathbf{p(x)} = \sigma(\small{\frac{1}{T} \frac{\mathbf{W}^{\mathrm{T}}F(\mathbf{x})}{\|F(\mathbf{x})\|}})$, where $\sigma{}$ indicates a softmax function.  In order to classify examples correctly, the direction of a weight vector has to be representative to the normalized features of the corresponding class. In this respect, the weight vectors can be regarded as estimated prototypes for each class. An architecture of our method is shown in Fig.~\ref{fig:pipeline}.

\subsection{Training Objectives}
\vspace{-1mm}
We estimate domain-invariant prototypes by performing entropy maximization with respect to the estimated prototype. Then, we extract discriminative features by performing entropy minimization with respect to feature extractor. Entropy maximization prevents overfitting that can reduce the expressive power of the representations. Therefore, entropy maximization can be considered as the step of selecting prototypes that will not cause overfitting to the source examples.
In our method, the prototypes are parameterized by the weight vectors of the last linear layer. 
First, we train $F$ and $C$ to classify labeled source and target examples correctly and utilize an entropy minimization objective to extract discriminative features for the target domain. 
We use a standard cross-entropy loss to train $F$ and $C$ for classification: 
\begin{equation}
\mathcal{L} = \mathbb{E}_{(\mathbf{x}, y) \in \mathcal{D}_{s}, \mathcal{D}_{t}} \mathcal{L}_{ce} \left( \mathbf{p(x)}, y \right).
\label{ce}
\end{equation}
With this classification loss, we ensure that the feature extractor generates discriminative features with respect to the source and a few target labeled examples. However, the model is trained on the source domain and a small fraction of target examples for classification. This does not learn discriminative features for the entire target domain. Therefore, we propose minimax entropy training using unlabeled target examples.

A conceptual overview of our proposed adversarial learning is illustrated in Fig.~\ref{fig:entropy-minmax}.
 We assume that there exists a single domain-invariant prototype for each class, which can be a representative point for both domains.
The estimated prototype will be near source distributions because source labels are dominant. Then, we propose to estimate the position of the prototype by moving each $\mathbf{w_i}$ toward target features using unlabeled data in the target domain.
To achieve this, we increase the entropy measured by the similarity between $\mathbf{W}$ and unlabeled target features. Entropy is calculated as follows, 
\vspace{-3mm}
\begin{equation}
H = - \mathbb{E}_{(\mathbf{x}, y) \in \mathcal{D}_{u}} \sum_{i=1}^{K} p(y=i|\mathbf{x}) \log p(y=i|\mathbf{x})
\end{equation}
where K is the number of classes and $p(y=i|\mathbf{x})$ represents the probability of prediction to class $i$, namely $i$ th dimension of $\mathbf{p(x)} = \sigma(\small{\frac{1}{T} \frac{\mathbf{W}^{\mathrm{T}}F(\mathbf{x})}{\|F(\mathbf{x})\|}})$.
To have higher entropy, that is, to have uniform output probability, each $\mathbf{w_i}$ should be similar to all target features.
Thus, increasing the entropy encourages the model to estimate the domain-invariant prototypes as shown in Fig.~\ref{fig:entropy-minmax}.

To obtain discriminative features on unlabeled target examples, we need to cluster unlabeled target features around the estimated prototypes. We propose to decrease the entropy on unlabeled target examples by the feature extractor $F$. The features should be assigned to one of the prototypes to decrease the entropy, resulting in the desired discriminative features. Repeating this prototype estimation (entropy maximization) and entropy minimization process yields discriminative features.

To summarize, our method can be formulated as  adversarial learning between $C$ and $F$. The task classifier $C$ is trained to maximize the entropy, whereas the feature extractor $F$ is trained to minimize it. Both $C$ and $F$ are also trained to classify labeled examples correctly.
The overall adversarial learning objective functions are:
\begin{equation}\label{eq:obj}
\begin{aligned}
\hat{\theta}_{F} &= \operatorname*{argmin}_{\theta_{F}} \mathcal{L} + \lambda H \\
\hat{\theta}_{C} &= \operatorname*{argmin}_{\theta_{C}} \mathcal{L} - \lambda H
\end{aligned}
\vspace{-1.5mm}
\end{equation}
where $\lambda$ is a hyper-parameter to control a trade-off between minimax entropy training and classification on labeled examples. Our method can be formulated as the iterative minimax training. To simplify training process, we use a gradient reversal layer~\cite{ganin2014unsupervised} to flip the gradient between $C$ and $F$ with respect to $H$. With this layer, we can perform the minimax training with one forward and back-propagation, which is illustrated in Fig.~\ref{fig:pipeline}.


\subsection{Theoretical Insights} \label{sec:theory}
\vspace{-1mm}
As shown in~\cite{ben2007analysis}, we can measure domain-divergence by using a domain classifier. 
Let $h \in \mathcal{H}$ be a hypothesis, ${\epsilon _s}(h)$ and ${\epsilon _t}(h)$ be the expected risk of source and target respectively, then
${\epsilon _t}(h ) \leqslant {\epsilon _s}(h ) + {d_\mathcal{H}}(p,q) + {C_0}$  where $C_0$ is a constant for the complexity of hypothesis space and the risk of an ideal hypothesis for both domains and ${d_\mathcal{H}}(p,q)$ is the ${\cal H}$-divergence between $p$ and $q$.
\vspace{-2mm}
\begin{equation}\label{eq:h_div}
	{d_\mathcal{H}}(p,q) \triangleq 2\mathop {\sup }\limits_{h  \in \mathcal{H}} \left| {\mathop {\Pr }\limits_{{{\mathbf{x}}^s} \sim p} \left[ {h ({{\mathbf{f}}^s}) = 1} \right] - \mathop {\Pr }\limits_{{{\mathbf{x}}^t} \sim q} \left[ {h({{\mathbf{f}}^t}) = 1} \right]} \right|
\end{equation}
where ${\mathbf{f}}^s$ and ${\mathbf{f}}^t$ denote the features in the source and target domain respectively. In our case the features are outputs of the feature extractor.
The ${\cal H}$-divergence relies on the capacity of the hypothesis space ${\cal H}$ to distinguish distributions $p$ and $q$. This theory states that the divergence between domains can be measured by training a domain classifier and features with low divergence are the key to having a well-performing task-specific classifier. Inspired by this, many methods~\cite{ganin2014unsupervised,bousmalis2016domain,tzeng2014deep,tzeng2017adversarial} train a domain classifier to discriminate different domains while also optimizing the feature extractor to minimize the divergence.

Our proposed method is also connected to Eq. \ref{eq:h_div}.
Although we do not have a domain classifier or a domain classification loss, our method can be considered as minimizing domain-divergence through minimax training on unlabeled target examples.  
We choose $h$ to be a classifier that decides a binary domain label of a feature by the value of the entropy, namely, 
\vspace{-1mm}
\begin{equation}
h(\mathbf{f}) = 
\begin{cases}
1, & \text{if } H(C(\mathbf{f})) \geq \gamma,\\
0, & \text{otherwise }
\end{cases}
\end{equation}
where $C$ denotes our classifier, $H$ denotes entropy, and $\gamma$ is a threshold to determine a domain label. Here, we assume $C$ outputs the probability of the class prediction for simplicity.
Eq.~\ref{eq:h_div} can be rewritten as follows, 
\vspace{1mm}
\scalebox{0.9}{
\begin{math}\label{eqn:Hdiverence}
	\begin{aligned}
	{d_\mathcal{H}}(p,q)& \triangleq 2\mathop {\sup }\limits_{h  \in \mathcal{H}} \left| {\mathop {\Pr }\limits_{{{\mathbf{f}}^s} \sim p} \left[ {h ({{\mathbf{f}}^s}) = 1} \right] - \mathop {\Pr }\limits_{{{\mathbf{f}}^t} \sim q} \left[ {h({{\mathbf{f}}^t}) = 1} \right]} \right|\\
	& =\textstyle{ 2\mathop {\sup }\limits_{C  \in \mathcal{C}}\left| \mathop  {\Pr }\limits_{{{\mathbf{f}}^s} \sim p} \left[ {H(C({{\mathbf{f}}^s})) \geq \gamma} \right] -  \mathop  {\Pr }\limits_{{{\mathbf{f}}^t} \sim q} \left[ {H(C({{\mathbf{f}}^t})) \geq \gamma} \right] \right|}\\
	& \leq  2\mathop {\sup }\limits_{C  \in \mathcal{C}}  \mathop  {\Pr }\limits_{{{\mathbf{f}}^t} \sim q} \left[ {H(C({{\mathbf{f}}^t})) \geq \gamma} \right].
	\end{aligned}
	\end{math}}
In the last inequality, we assume that  {\small $\mathop  {\Pr }\limits_{{{\mathbf{f}}^s} \sim p} \left[ {H(C({{\mathbf{f}}^s})) \geq \gamma} \right] \leq  \mathop  {\Pr }\limits_{{{\mathbf{f}}^t} \sim p} \left[ {H(C({{\mathbf{f}}^t})) \geq \gamma} \right]$}. This assumption should be realistic because we have access to many labeled source examples and train entire networks to minimize the classification loss. Minimizing the cross-entropy loss (Eq.~\ref{ce}) on source examples ensures that the entropy on a source example is very small.
Intuitively, this inequality states that the divergence can be bounded by the ratio of target examples having entropy greater than $\gamma$. Therefore, we can have the upper bound by finding the $C$ that achieves maximum entropy for all target features. Our objective is finding features that achieve lowest divergence. We suppose there exists a $C$ that achieves the maximum in the inequality above, then the objective can be rewritten as, 
\vspace{-1mm}
\begin{equation}
\min_{\mathbf{f}^t}\mathop{\max }\limits_{C  \in \mathcal{C}}  \mathop  {\Pr }\limits_{{{\mathbf{f}}^t} \sim q} \left[ {H(C({{\mathbf{f}}^t})) \geq \gamma} \right]
\end{equation}
Finding the minimum with respect to ${\mathbf{f}}^t$ is equivalent to find a feature extractor $F$ that achieves that minimum. 
Thus, we derive the minimax objective of our proposed learning method in Eq .~\ref{eq:obj}.  To sum up, our maximum entropy process can be regarded as measuring the divergence between domains, whereas our entropy minimization process can be regarded as minimizing the divergence. In our experimental section, we observe that our method actually reduces domain-divergence (Fig. \ref{fig:a-distance}). In addition, target features produced by our method look aligned with source features and are just as discriminative. These come from the effect of the domain-divergence minimization. 

%% file: experiment.tex
\renewcommand{\arraystretch}{1.1}

\begin{table*}[t]
\begin{center}
\scalebox{0.85}{
\begin{tabular}{l|l|cccccccccccccc|cc}
\toprule[1.5pt] 
 \multirow{2}{*}{Net} & \multirow{2}{*}{Method}       &\multicolumn{2}{c}{R to C}&\multicolumn{2}{c}{R to P} & \multicolumn{2}{c}{P to C}  & \multicolumn{2}{c}{C to S} & \multicolumn{2}{c}{S to P} & \multicolumn{2}{c}{R to S} & \multicolumn{2}{c}{P to R}     &\multicolumn{2}{|c}{MEAN} \\ 
& &1\scriptsize{-shot}&3\scriptsize{-shot} &1\scriptsize{-shot}&3\scriptsize{-shot}&1\scriptsize{-shot}&3\scriptsize{-shot} &1\scriptsize{-shot}&3\scriptsize{-shot}&1\scriptsize{-shot}&3\scriptsize{-shot}&1\scriptsize{-shot}&3\scriptsize{-shot} &1\scriptsize{-shot}&3\scriptsize{-shot} &1\scriptsize{-shot}&3\scriptsize{-shot}  \\ \hline
 \multirow{6}{*}{AlexNet} &S+T& 43.3   & 47.1 & 42.4   & 45.0 & 40.1   & 44.9 & 33.6   & 36.4 & 35.7   & 38.4 & 29.1 & 33.3 & 55.8   & 58.7 & 40.0 & 43.4 \\
&DANN & 43.3   & 46.1 & 41.6   & 43.8 & 39.1   & 41.0 & 35.9   & 36.5 &36.9   & 38.9 & 32.5 & 33.4 & 53.6   & 57.3 & 40.4 & 42.4 \\
&ADR & 43.1        & 46.2 &    41.4    & 44.4 &    39.3    & 43.6 & 32.8        &   36.4   &  33.1      &  38.9    &   29.1   &  32.4    & 55.9  & 57.3 & 39.2  & 42.7 \\
&CDAN& 46.3   & 46.8 & 45.7   & 45.0 & 38.3   & 42.3 & 27.5   & 29.5 & 30.2   & 33.7 & 28.8 & 31.3 & 56.7   & 58.7 & 39.1 & 41.0 \\
&ENT          & 37.0   & 45.5 & 35.6   & 42.6 & 26.8   & 40.4 & 18.9   & 31.1 & 15.1   & 29.6 & 18.0 & 29.6 & 52.2   & 60.0 & 29.1 & 39.8 \\
&MME & \bf{48.9}& \bf{55.6} & \bf{48.0}   & \bf{49.0} & \bf{46.7}   &\bf{51.7} & \bf{36.3}   & \bf{39.4} & \bf{39.4}   & \bf{43.0} & \bf{33.3} & \bf{37.9} & \bf{56.8}   & \bf{60.7} & \bf{44.2} & \bf{48.2}\\\hline\hline
\multirow{6}{*}{VGG}      & S+T          & 49.0 & 52.3   & 55.4 & 56.7   & 47.7 & 51.0   & 43.9 & 48.5         & 50.8 & 55.1         & 37.9 & 45.0   & 69.0           & 71.7        & 50.5 & 54.3 \\
         & DANN         & 43.9 & 56.8   & 42.0 & 57.5   & 37.3 & 49.2   & 46.7 &48.2   & 51.9 &55.6 & 30.2 & 45.6   & 65.8           & 70.1     & 45.4 & 54.7 \\
         & ADR          &   48.3   & 50.2   &   54.6   & 56.1   &  47.3    & 51.5   &     44.0  &   49.0   & 50.7    & 53.5     &  38.6      &  44.7    & 67.6         & 70.9 &   50.2 &53.7 \\
         & CDAN         & 57.8 & 58.1   & 57.8 & 59.1   & 51.0 & 57.4   & 42.5 & 47.2         & 51.2 & 54.5         & 42.6 & 49.3   & 71.7           & 74.6              &53.5  & 57.2 \\
         & ENT          & 39.6 & 50.3   & 43.9 & 54.6   & 26.4 & 47.4   & 27.0 & 41.9         & 29.1 & 51.0         & 19.3 & 39.7   & 68.2           & 72.5           & 36.2 & 51.1 \\
         & MME & \bf{60.6} & \bf{64.1}   & \bf{63.3} & \bf{63.5}   & \bf{57.0} & \bf{60.7}   &\bf{50.9} & \bf{55.4}         & \bf{60.5} & \bf{60.9}         & \bf{50.2} & \bf{54.8}   & \bf{72.2}           & \bf{75.3}                  & \bf{59.2} & \bf{62.1} \\\hline\hline
\multirow{6}{*}{ResNet} &S+T    & 55.6 & 60.0   & 60.6 & 62.2   & 56.8 & 59.4   & 50.8 & 55.0   & 56.0 & 59.5 & 46.3 & 50.1   & 71.8 & 73.9 & 56.9 & 60.0 \\
&DANN   & 58.2 & 59.8   & 61.4 & 62.8   & 56.3 & 59.6   & 52.8 & 55.4   & 57.4 & 59.9 & 52.2 & 54.9   & 70.3 & 72.2 & 58.4 & 60.7 \\
&ADR    & 57.1 & 60.7 & 61.3 & 61.9 & 57.0 & 60.7 & 51.0 & 54.4 & 56.0 & 59.9 & 49.0 & 51.1 & 72.0 & 74.2 & 57.6 & 60.4
  \\
&CDAN   & 65.0 & 69.0   & 64.9 & 67.3   & 63.7 & 68.4   & 53.1 & 57.8   & 63.4 & 65.3 & 54.5 & 59.0   & 73.2 & 78.5 & 62.5 & 66.5 \\
&ENT    & 65.2 & 71.0   & 65.9 & 69.2   & 65.4 & 71.1   & 54.6 & 60.0   & 59.7 & 62.1 & 52.1 & 61.1   & 75.0 & \bf{78.6} & 62.6 & 67.6 \\
&MME   & \bf{70.0} & \bf{72.2}   & \bf{67.7} & \bf{69.7}   & \bf{69.0} & \bf{71.7}   & \bf{56.3} & \bf{61.8}   & \bf{64.8} & \bf{66.8} & \bf{61.0} & \bf{61.9}   & \bf{76.1} & 78.5 & \bf{66.4} & \bf{68.9}\\
        \bottomrule[1.5pt]
\end{tabular}}
\end{center}
\vspace{-5mm}
\caption{Accuracy on the DomainNet dataset (\%) for one-shot and three-shot settings on 4 domains, R: Real, C: Clipart, P: Clipart, S: Sketch. Our MME method outperformed other baselines for all adaptation scenarios and for all three networks, except for only one case where it performs similarly to ENT.}
\label{tb:lsdac_all}

\end{table*}

\section{Experiments}
\vspace{-1mm}
\subsection{Setup}
\vspace{-1mm}
We randomly selected one or three labeled examples per class as the labeled training target examples (one-shot and three-shot setting, respectively.) We selected three other labeled examples as the validation set for the target domain. The validation examples are used for early stopping, choosing the hyper-parameter $\lambda$, and training scheduling. The other target examples are used for training without labels, their labels are only used to evaluate classification accuracy (\%). All examples of the source are used for training. 

\noindent
\textbf{Datasets.} 
Most of our experiments are done on a subset of \textbf{DomainNet}~\cite{peng2018moment}, a recent benchmark dataset for large-scale domain adaptation that has many classes (345) and six domains. 
As labels of some domains and classes are very noisy, we pick 4 domains (Real, Clipart, Painting, Sketch) and 126 classes. 
We focus on the adaptation scenarios where the target domain is not real images, and construct 7 scenarios from the four domains. See our supplemental material for more details. 
\textbf{Office-Home}~\cite{venkateswara2017Deep} contains 4 domains (Real, Clipart, Art, Product) with 65 classes. This dataset is one of the benchmark datasets for unsupervised domain adaptation. We evaluated our method on 12 scenarios in total.
\textbf{Office}~\cite{saenko2010} contains 3 domains (Amazon, Webcam, DSLR) with 31 classes. Webcam and DSLR are small domains and some classes do not have a lot of examples while Amazon has many examples. 
To evaluate on the domain with enough examples, we have 2 scenarios where we set Amazon as the target domain and DSLR and Webcam as the source domain.  

\noindent
\textbf{Implementation Details.}
All experiments are implemented in Pytorch~\cite{paszke2017automatic}. 
We employ AlexNet~\cite{krizhevsky2012imagenet} and VGG16~\cite{simonyan2014very} pre-trained on ImageNet. To investigate the effect of deeper architectures, we use ResNet34~\cite{he2016deep} in experiments on DomainNet. 
We remove the last linear layer of these networks to build $F$, and add a K-way linear classification layer $C$ with a randomly initialized weight matrix $W$. The value of temperature $T$ is set 0.05 following the results of~\cite{ranjan2017l2} in all settings. 
Every iteration, we prepared two mini-batches, one consisting of labeled examples and the other of unlabeled target examples. Half of the labeled examples comes from source and half from labeled target. Using the two mini-batches, we calculated the objective in Eq.~\ref{eq:obj}. 
To implement the adversarial learning in Eq.~\ref{eq:obj}, we use a gradient reversal layer~\cite{ganin2014unsupervised,tzeng2014deep} to flip the gradient with respect to entropy loss. The sign of the gradient is flipped between $C$ and $F$ during backpropagation. 
We adopt SGD with momentum of 0.9. In all experiments, we set the trade-off parameter $\lambda$ in Eq. \ref{eq:obj} as $0.1$. This is decided by the validation performance on Real to Clipart experiments. We show the performance sensitivity to this parameter in our supplemental material, as well as more details including learning rate scheduling. 

\begin{table}[t]
\begin{center}
\scalebox{0.9}{
\begin{tabular}{l|l|cccc}
\toprule[1.5pt] 
 \multirow{2}{*}{Net} & \multirow{2}{*}{Method}       &\multicolumn{2}{c}{Office-Home}&\multicolumn{2}{c}{Office} \\
 &        & 1-shot & 3-shot & 1-shot & 3-shot  \\ \hline
\multirow{6}{*}{AlexNet}  & S+T          & 44.1     & 50.0  &50.2& 61.8    \\
        & DANN         & 45.1     & 50.3&55.8&64.8       \\
        & ADR          & 44.5 & 49.5      &50.6&61.3 \\
        & CDAN         & 41.2     & 46.2    &49.4&60.8   \\
         & ENT          & 38.8     & 50.9     &48.1&65.1  \\
        & MME & \bf{49.2}     & \bf{55.2}     &\bf{56.5}&\bf{67.6}  \\\hline\hline

\multirow{6}{*}{VGG}        & S+T          & 57.4     & 62.9&68.7 & 73.3 \\
        & DANN         & 60.0     & 63.9& 69.8 &    75.0 \\
        & ADR          & 57.4 & 63.0& 69.4 & 73.7       \\
        & CDAN         & 55.8     & 61.8& 65.9 &  72.9 \\
        & ENT          & 51.6     & 64.8 &70.6  & 75.3     \\
        & MME & \bf{62.7}     & \bf{67.6}& \bf{73.4} & \bf{77.0}   \\
         \bottomrule[1.5pt]
\end{tabular}}
\end{center}
\vspace{-5mm}
\caption{Results on Office-Home and Office dataset (\%). The value is the accuracy averaged over all adaptation scenarios. Performance on each setting is summarized in supplementary material.}
\label{tb:tb_office_home}

\end{table}
\noindent
\textbf{Baselines.} 
\textbf{S+T}~\cite{chen2018closer,ranjan2017l2} is a model trained with the labeled source and labeled target examples without using unlabeled target examples. 
\textbf{DANN~\cite{ganin2014unsupervised}} employs a domain classifier to match feature distributions. This is one of the most popular methods in UDA. For fair comparison, we modify this method so that it is trained with the labeled source, labeled target, and unlabeled target examples.  
\textbf{ADR~\cite{ADR}} utilizes a task-specific decision boundary to align features and ensure that they are discriminative on the target. 
\textbf{CDAN~\cite{long2017conditional}} is one of the state-of-the art methods on UDA and performs domain alignment on features that are conditioned on the output of classifiers. In addition, it utilizes entropy minimization on target examples. CDAN integrates domain-classifier based alignment and entropy minimization.
Comparison with these UDA methods (DANN, ADR, CDAN) reveals how much gain will be obtained compared to the existing domain alignment-based methods.
 \textbf{ENT~\cite{grandvalet2005semi}} is a model trained with labeled source and target and unlabeled target using standard entropy minimization. Entropy is calculated on  unlabeled target examples and the entire network is trained to minimize it. The difference from MME is that ENT does not have a maximization process, thus comparison with this baseline clarifies its importance.
 
Note that all methods except for CDAN are trained with exactly the same architecture used in our method. In case of CDAN, we could not find any advantage of using our architecture. The details of baseline implementations are in our supplemental material.

\subsection{Results}
    \vspace{-1mm}
\noindent
\textbf{Overview}.
The main results on the DomainNet dataset are shown in Table \ref{tb:lsdac_all}. First, our method outperformed other baselines for all adaptation scenarios and all three networks except for one case. 
On average, our method outperformed S+T with 9.5\% and 8.9\% in ResNet one-shot and three-shot setting respectively. The results on Office-Home and Office are summarized in Table \ref{tb:tb_office_home}, where MME also outperforms all baselines. Due to the limited space, we show the results averaged on all adaptation scenarios. 

\noindent
\textbf{Comparison with UDA Methods}.
Generally, baseline UDA methods need strong base networks such as VGG or ResNet to perform better than S+T. Interestingly, these methods cannot improve the performance in some cases. The superiority of MME over existing UDA methods is supported by Tables \ref{tb:lsdac_all} and \ref{tb:tb_office_home}.
Since CDAN uses entropy minimization and ENT significantly hurts the performance for AlexNet and VGG, CDAN does not consistently improve the performance for AlexNet and VGG.

\noindent
\textbf{Comparison with Entropy Minimization}.
ENT does not improve performance in some cases because it does not account for the domain gap. Comparing results on one-shot and three-shot, entropy minimization gains performance with the help of labeled examples. As we have more labeled target examples, the estimation of prototypes will be more accurate without any adaptation. In case of ResNet, entropy minimization often improves accuracy. There are two potential reasons. First, ResNet pre-trained on ImageNet has a more discriminative representation than other networks. Therefore, given a few labeled target examples, the model can extract more discriminative features, which contributes to the performance gain in entropy minimization. Second, ResNet has batch-normalization (BN) layers~\cite{ioffe2015batch}. It is reported that BN has the effect of aligning feature distributions~\cite{cariucci2017autodial,li2016revisiting}. Hence, entropy minimization was done on aligned feature representations, which improved the performance. When there is a large domain gap such as C to S, S to P, and R to S in Table \ref{tb:lsdac_all}, BN is not enough to handle the domain gap. Therefore, our proposed method performs much better than entropy minimization in such cases. We show an analysis of BN in our supplemental material, revealing its effectiveness for entropy minimization. 
\begin{table}[]
\begin{center}
\scalebox{0.8}{
\begin{tabular}{l|ccccccc|c}
\toprule[1.5pt]
Method       & R - C & R - P & P - C & C - S & S - P & R - S & P - R & Avg \\\hline
Source            & 41.1   & 42.6   & 37.4   & 30.6   & 30.0   & 26.3   & 52.3   & 37.2 \\
DANN         & 44.7   & 36.1   & 35.8   & 33.8   & \bf{35.9}   & 27.6   & 49.3   & 37.6 \\
ADR          & 40.2   & 40.1   & 36.7   & 29.9   & 30.6   & 25.9   & 51.5   & 36.4 \\
CDAN         & 44.2   & 39.1   & 37.8   & 26.2   & 24.8   & 24.3   & \bf{54.6}   & 35.9 \\
ENT          & 33.8   & 43.0   & 23.0   & 22.9   & 13.9   & 12.0   & 51.2   & 28.5 \\
MME & \bf{47.6}   & \bf{44.7}   & \bf{39.9}   & \bf{34.0}   & 33.0   & \bf{29.0}   & 53.5   & \bf{40.2} \\
\bottomrule[1.5pt]
\end{tabular}
}
\end{center}
\vspace{-5mm}
\caption{Results on the DomainNet dataset in the unsupervised domain adaptation setting (\%).}
    \label{tb:lsda_uda}

\end{table}

\begin{figure}[tbp]
 \begin{subfigure}{0.49\hsize}
  \begin{center}
   \includegraphics[width=1.05\hsize]{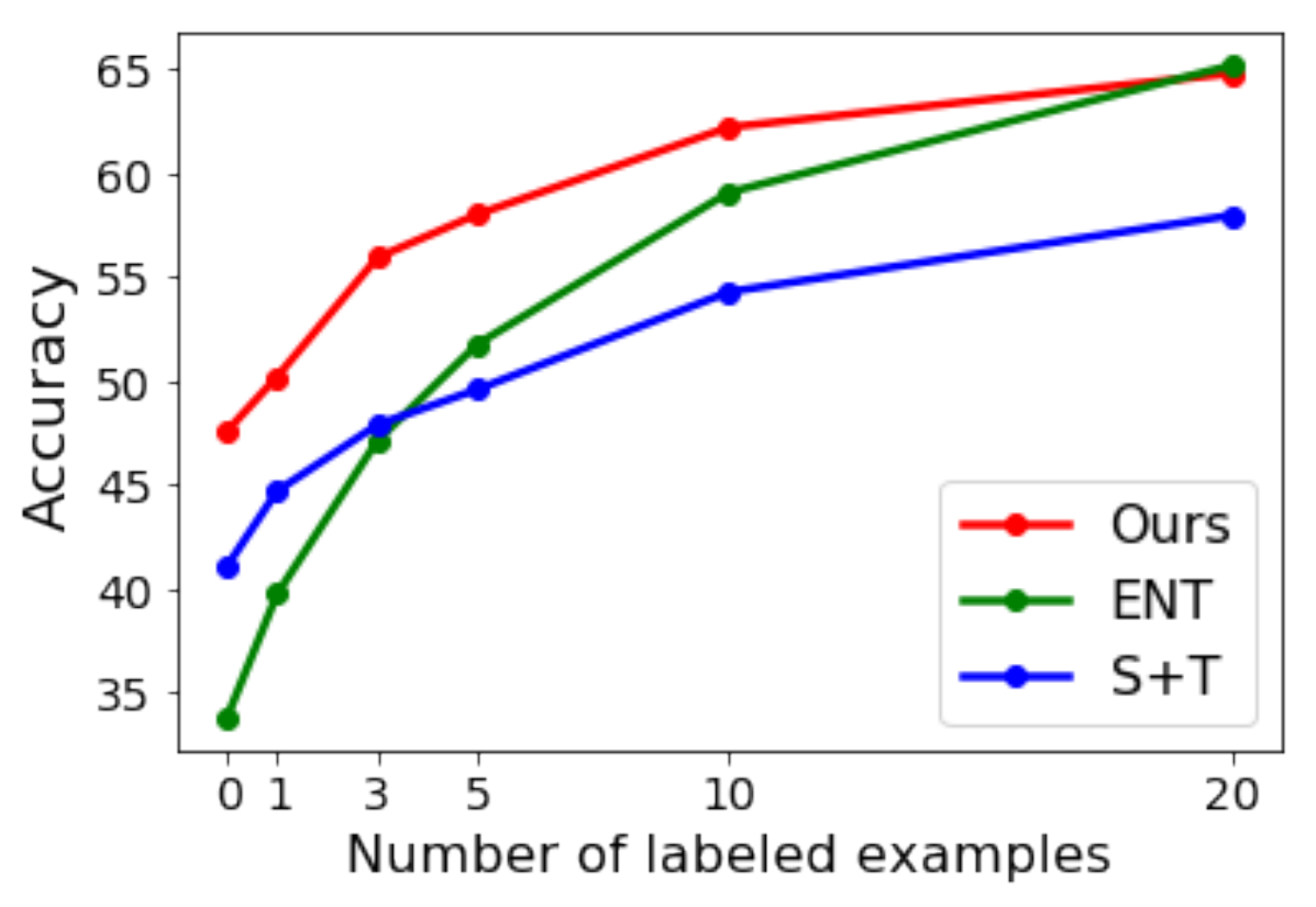}
   \caption{AlexNet}
  \end{center}
  
 \end{subfigure}
 \begin{subfigure}{0.49\hsize}
  \begin{center}
   \includegraphics[width=1.05\hsize]{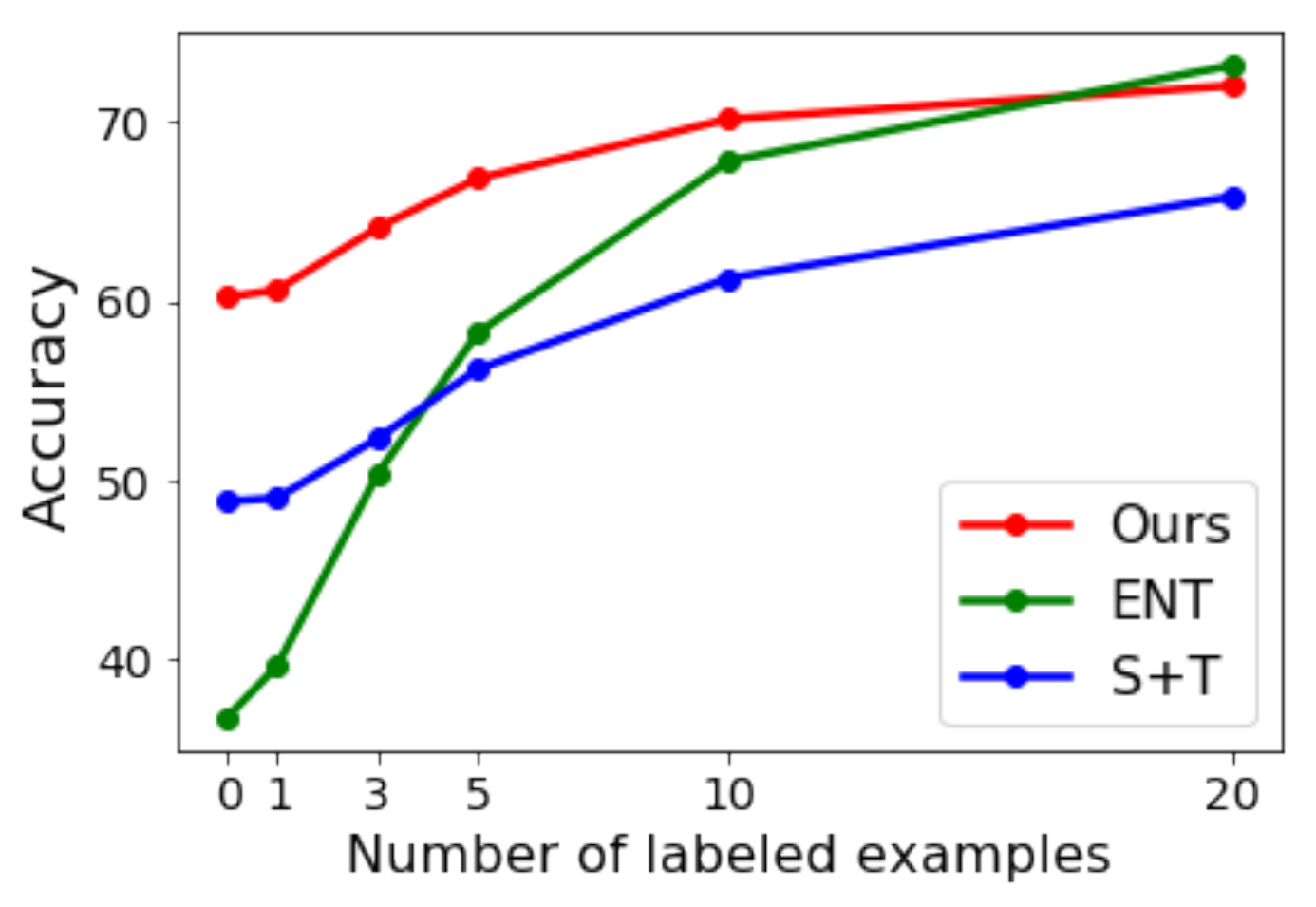}
   \caption{VGG}
  \end{center}
 
 \end{subfigure}
 \vspace{-3mm}
  \caption{Accuracy vs the number of labeled target examples. The ENT method needs more labeled examples to obtain similar performance to our method.}
   \label{fig:change_num}
\end{figure}
\vspace{-2mm}
\begin{table}[t]
\begin{center}
\scalebox{0.8}{
\begin{tabular}{l|cccc}
\toprule[1.5pt] 
 \multirow{2}{*}{Method}       &\multicolumn{2}{c}{R to C} &\multicolumn{2}{c}{R to S}\\
         & 1-shot & 3-shot & 1-shot & 3-shot  \\ \hline
         S+T (Standard Linear)          & 41.4 &44.3 &26.5&28.7   \\
         S+T (Few-shot~\cite{chen2018closer,ranjan2017l2})          & \bf{43.3} &\bf{47.1}&\bf{29.1}&\bf{33.3}    \\\hline
         MME (Standard Linear)          &44.9  &47.7   &30.0&32.2 \\
         MME (Few-shot~\cite{chen2018closer,ranjan2017l2})        & \bf{48.9} &\bf{55.6}&\bf{33.3}&\bf{37.9}    \\
         \bottomrule[1.5pt]
\end{tabular}}
\end{center}
\vspace{-5mm}
\caption{Comparison of classifier architectures on the DomainNet dataset using AlexNet, showing the effectiveness of the architecture proposed in~\cite{chen2018closer,ranjan2017l2}.}
\label{tb:tb_ablation}

\end{table}

\begin{figure*}[t]
    \centering
    \begin{subfigure}[b]{0.23\textwidth}
        \includegraphics[width=\textwidth]{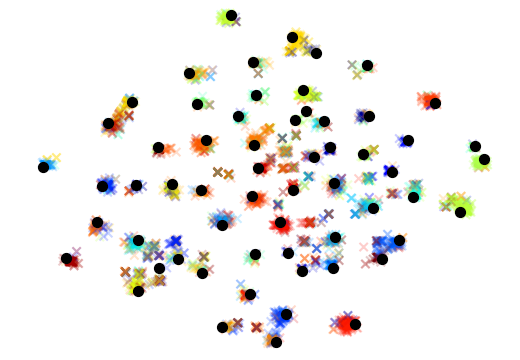}
        \caption{Ours}
    \end{subfigure}
    ~ 
    \begin{subfigure}[b]{0.23\textwidth}
        \includegraphics[width=\textwidth]{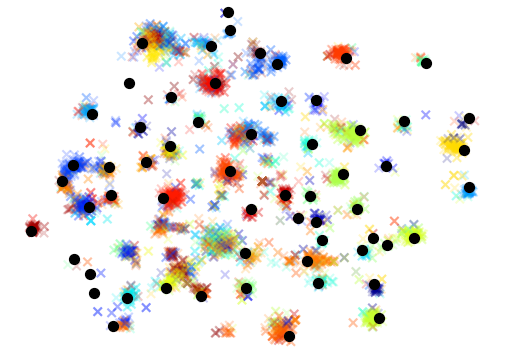}
        \caption{ENT}
        
    \end{subfigure}
    \begin{subfigure}[b]{0.23\textwidth}
        \includegraphics[width=\textwidth]{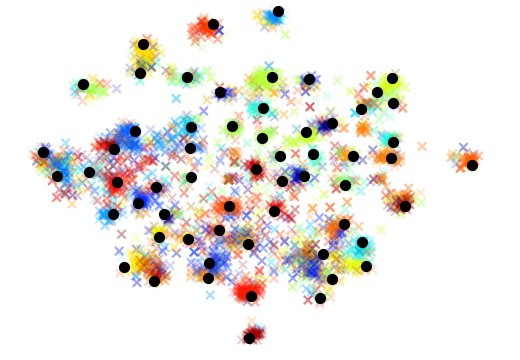}
        \caption{DANN}
    \end{subfigure}
    ~ 
    \begin{subfigure}[b]{0.23\textwidth}
        \includegraphics[width=\textwidth]{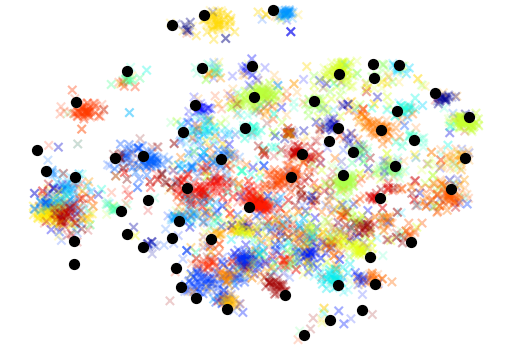}
        \caption{S+T}
    \end{subfigure}
    
    \begin{subfigure}[b]{0.23\textwidth}
        \includegraphics[width=\textwidth]{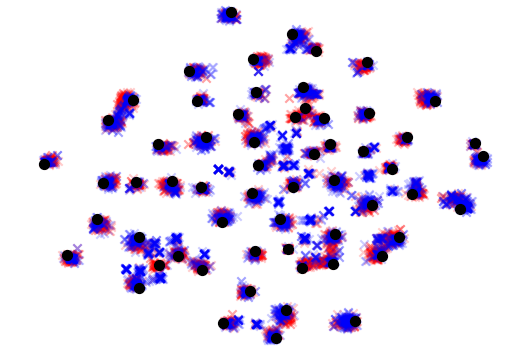}
        \caption{Ours}
        \label{fig:gull}
    \end{subfigure}
    ~ 
    \begin{subfigure}[b]{0.23\textwidth}
        \includegraphics[width=\textwidth]{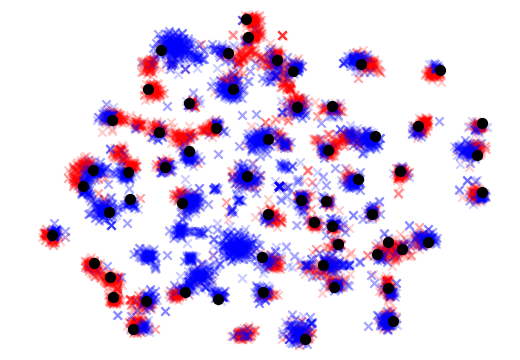}
        \caption{ENT}
         \label{fig:vis_domain_ent}
    \end{subfigure}
    \begin{subfigure}[b]{0.23\textwidth}
        \includegraphics[width=\textwidth]{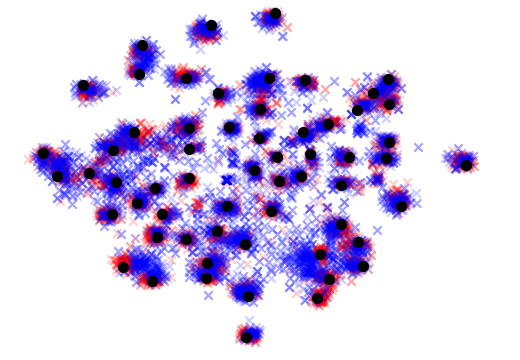}
        \caption{DANN}
    \end{subfigure}
    ~ 
    \begin{subfigure}[b]{0.23\textwidth}
        \includegraphics[width=\textwidth]{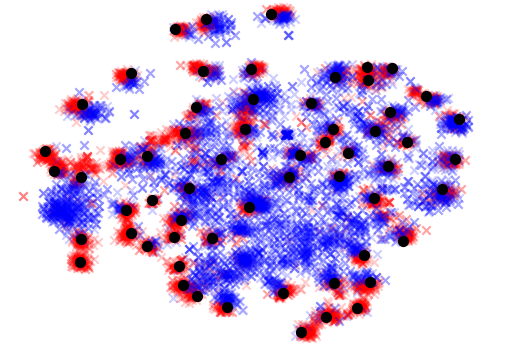}
        \caption{S+T}
    \end{subfigure}
    \vspace{-1mm}
    \caption{Feature visualization with t-SNE. (a-d) We plot the class prototypes (black circles) and features on the target domain (crosses). The color of a cross represents its class. We observed that features on our method show more discrimative features than other methods. (e-h) \textcolor{red}{Red}: Features of the source domain. \textcolor{blue}{Blue}: Features of the target domain. Our method's features are well-aligned between domains compared to other methods.
    }
    \label{fig:FeatureViz}
\end{figure*}

\begin{figure*}[h!]

\begin{subfigure}{0.31\hsize}
  \begin{center}
   \includegraphics[width=\hsize]{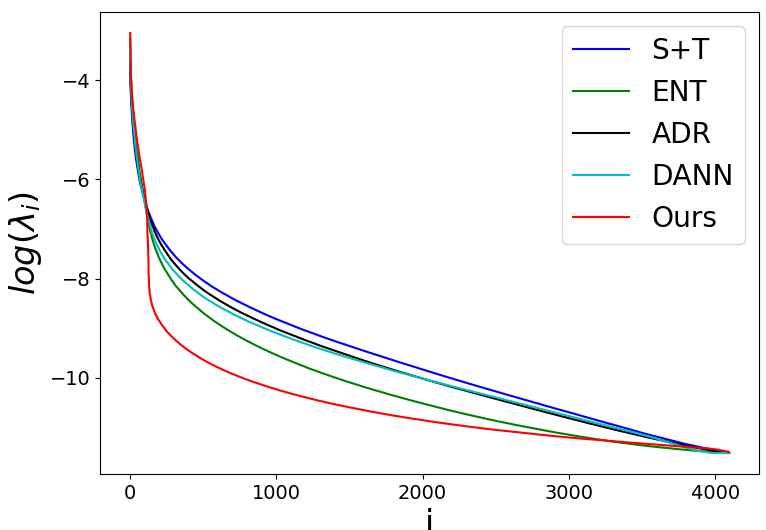}
   \caption{Eigenvalues}
   \label{fig:logi}
  \end{center}
  
 \end{subfigure}
 \begin{subfigure}{0.31\hsize}
  \begin{center}
   \includegraphics[width=\hsize]{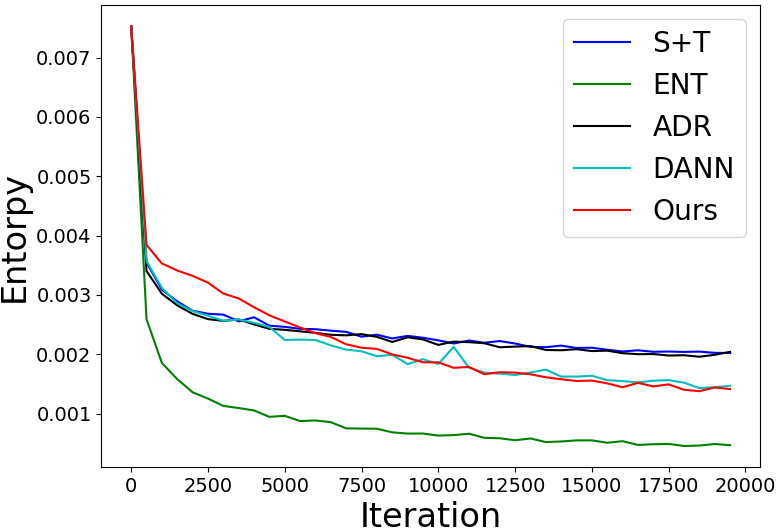}
   \caption{Entropy}
   \label{fig:entropy}
  \end{center}
 \end{subfigure}
  \begin{subfigure}{0.32\hsize}
  \begin{center}
   \includegraphics[width=\hsize]{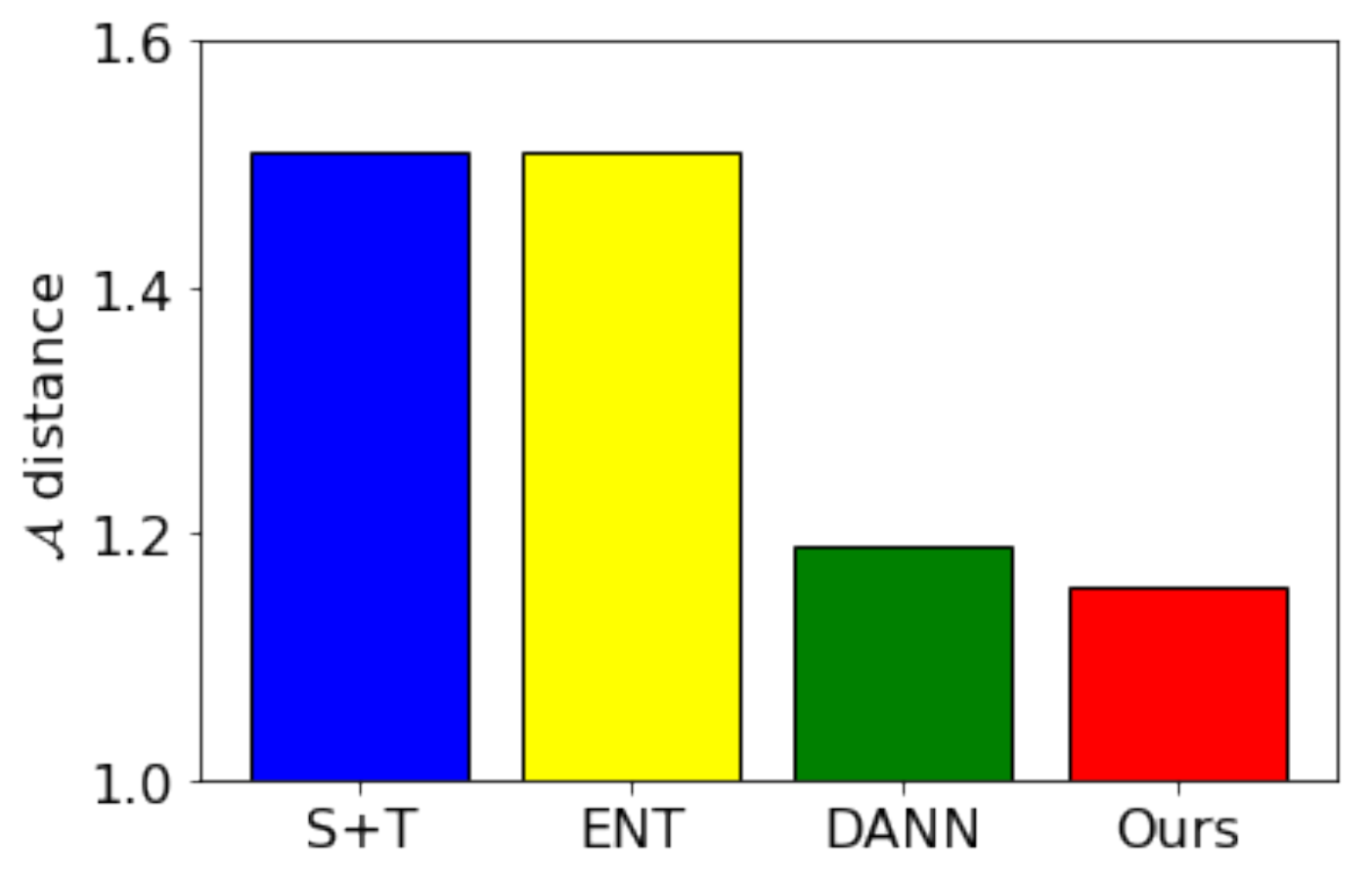}
   \caption{$\mathcal{A}$-distance}
   \label{fig:a-distance}
  \end{center}
 \end{subfigure}
 \vspace{-1mm}
    \caption{(a) Eigenvalues of the covariance matrix of the features on the target domain. Eigenvalues reduce quickly in our method, which shows that features are more discriminative than other methods.   (b) Our method achieves lower entropy than baselines except ENT. (c) Our method clearly reduces domain-divergence compared to S+T.}
    \label{fig:DiversityAnalysis}
\end{figure*}

\subsection{Analysis}
\vspace{-2mm}
\noindent
\textbf{Varying Number of Labeled Examples}.
First, we show the results on unsupervised domain adaptation setting in Table \ref{tb:lsda_uda}. 
Our method performed better than other methods on average. In addition, only our method improved performance compared to source only model in all settings. 
Furthermore, we observe the behavior of our method when the number of labeled examples in the target domain varies from 0 to 20 per class, which corresponds to 2520 labeled examples in total. The results are shown in Fig. \ref{fig:change_num}. Our method works much better than S+T given a few labeled examples. On the other hand, ENT needs 5 labeled examples per class to improve  performance. As we add more labeled examples, the performance gap between ENT and ours is reduced. This result is quite reasonable, because prototype estimation will become more accurate without any adaptation as we have more labeled target examples. 

\noindent
\textbf{Effect of Classifier Architecture}.
 We introduce an ablation study on the classifier network architecture proposed in~\cite{chen2018closer,ranjan2017l2} with AlexNet on DomainNet. As shown in Fig. \ref{fig:pipeline}, we employ $\ell_2$ normalization and temperature scaling. In this experiment, we compared it with a model having a standard linear layer without $\ell_2$ normalization and temperature. The result is shown in Table \ref{tb:tb_ablation}. By using the network architecture proposed in ~\cite{chen2018closer,ranjan2017l2}, we can improve the performance of both our method and the baseline S+T model (model trained only on source examples and a few labeled target examples.) Therefore, we can argue that the network architecture is an effective technique to improve performance when we are given a few labeled examples from the target domain.

\noindent
\textbf{Feature Visualization.}
In addition, we plot the learned features with t-SNE \cite{maaten2008visualizing} in Fig. \ref{fig:FeatureViz}. We employ the scenario Real to Clipart of DomainNet using AlexNet as the pre-trained backbone. Fig \ref{fig:FeatureViz} (a-d) visualizes the target features and estimated prototypes. The color of the cross represents its class, black points are the prototypes. 
With our method, the target features are clustered to their prototypes and do not have a large variance within the class. 
 We visualize features on the source domain (red cross) and target domain (blue cross) in Fig.~\ref{fig:FeatureViz} (e-h). As we discussed in the method section, our method aims to minimize domain-divergence. Indeed, target features are well-aligned with source features with our method. Judging from Fig.~\ref{fig:vis_domain_ent}, entropy minimization (ENT) also tries to extract discriminative features, but it fails to find domain-invariant prototypes.

\noindent
\textbf{Quantitative Feature Analysis.} 
We quantitatively investigate the characteristics of the features we obtain using the same adaptation scenario.
First, we perform the analysis on the eigenvalues of the covariance matrix of target features. We follow the analysis done in \cite{dubey2018maximum}. Eigenvectors represent the components of the features and eigenvalues represent their contributions. If the features are highly discriminative, only a few components are needed to summarize them. Therefore, in such a case, the first few eigenvalues are expected to be large, and the rest to be small. The features are clearly summarized by fewer components in our method as shown in Fig.~\ref{fig:logi}. Second, we show the change of entropy value on the target in Fig.~\ref{fig:entropy}. ENT diminishes the entropy quickly, but results in poor performance. This indicates that the method increases the confidence of predictions incorrectly while our method achieves higher accuracy at the same time. Finally, in Fig.~\ref{fig:a-distance}, we calculated $\mathcal{A}$-distance by training a SVM as a domain classifier as proposed in~\cite{ben2007analysis}. Our method greatly reduces the distance compared to S+T. The claim that our method reduces a domain divergence is empirically supported with this result.

%% file: conclusion.tex
\vspace{-2mm}
\section{Conclusion}
\vspace{-2mm}
We proposed a novel Minimax Entropy (MME) approach that adversarially optimizes an adaptive few-shot model for semi-supervised domain adaptation (SSDA). Our model consists of a feature encoding network, followed by a classification layer that computes the features' similarity to a set of estimated prototypes (representatives of each class). Adaptation is achieved by alternately maximizing the conditional entropy of unlabeled target data with respect to the classifier and minimizing it with respect to the feature encoder.
We empirically demonstrated the superiority of our method over many baselines, including conventional feature alignment and few-shot methods, setting a new state of the art for SSDA. 
\vspace{-2mm}
\section{Acknowledgements}
\vspace{-2mm}
This work was supported by Honda, DARPA, BAIR, BDD, and NSF Award No. 1535797.

%% file: supp_arxiv.tex
\setcounter{section}{0}
\vspace{20mm}
\section*{Supplemental Material}
\section{Datasets}
\begin{figure*}[t]
\begin{center}
   \includegraphics[width=\linewidth]{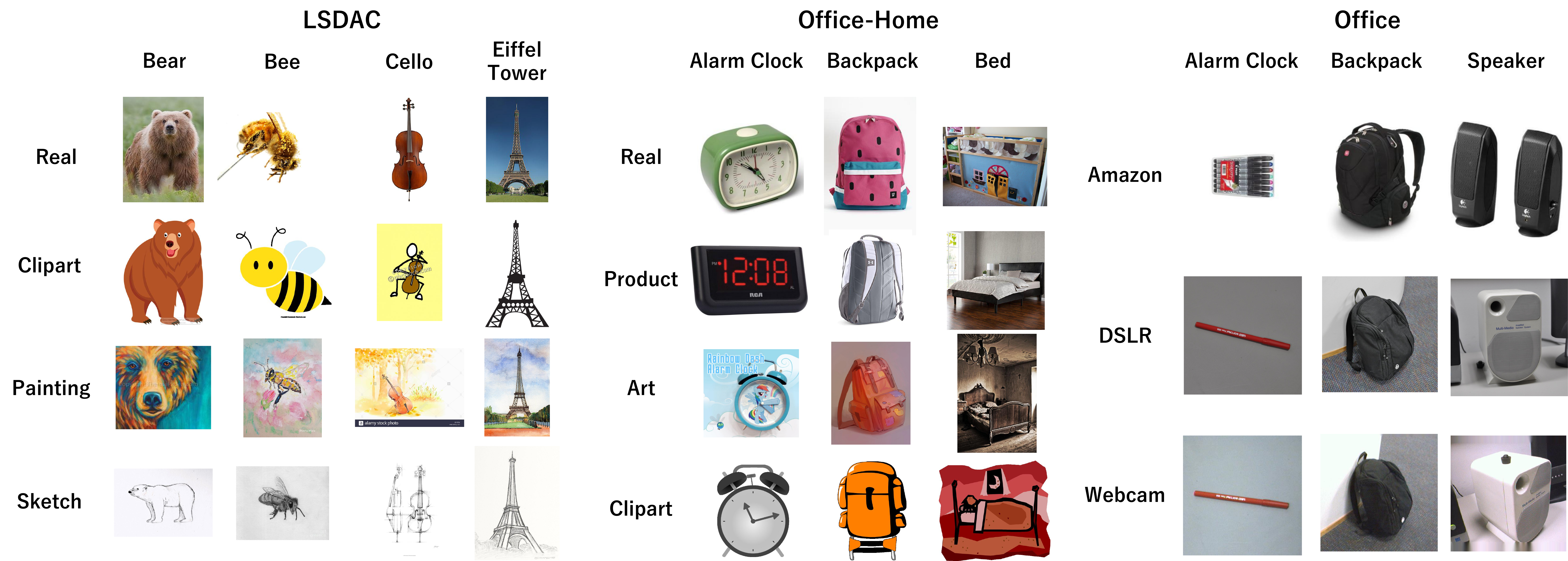}
\end{center}
\vspace{-5mm}
   \caption{Example images in DomainNet, Office-Home, and Office.}
\label{fig:dataset}
\end{figure*}

First, we show the examples of datasets we employ in the experiments in Fig~\ref{fig:dataset}. We also attach a list of classes used in our experiments on DomainNet with this material.
\section{Implementation Detail}
We provide details of our implementation. {\bf We will publish our implementation upon acceptance.}
The reported performance in the main paper is obtained by one-time training. In this material, we also report both average and variance on multiple runs and results on different dataset splits (i.e., different train/val split). 

\textbf{Implementation of MME.}
For VGG and AlexNet, we replace the last linear layer with randomly initialized linear layer. With regard to ResNet34, we remove the last linear layer and add two fully-connected layers following~\cite{long2017conditional}. 
We use the momentum optimizer where the initial learning rate is set 0.01 for all fully-connected layers whereas it is set 0.001 for other layers including convolution layers and batch-normalization layers. We employ learning rate annealing strategy proposed in~\cite{ganin2014unsupervised}. 
Each mini-batch consists of labeled source, labeled target and unlabeled target images. Labeled examples and unlabeled examples are separately forwarded. We sample $s$ labeled source and labeled target images and $2s$ unlabeled target images. $s$ is set to be 32 for AlexNet, but 24 for VGG and ResNet due to GPU memory contraints.
We use horizontal-flipping and random-cropping based data augmentation for all training images.

\subsection{Baseline Implementation}
Except for CDAN, we implemented all baselines by ourselves.
\textbf{S+T}~\cite{chen2018closer}. This approach only uses labeled source and target examples with the cross-entropy loss for training. 

\textbf{DANN}~\cite{ganin2014unsupervised}. We train a domain classifier on the output of the feature extractor. It has three fully-connected layers with relu activation. The dimension of the hidden layer is set 512. We use a sigmoid activation only for the final layer. The domain classifier is trained to distinguish source examples and unlabeled target examples. 

\textbf{ADR}~\cite{ADR}. We put dropout layer with 0.1 dropout rate after l2-normalization layer. For unlabeled target examples, we calculate sensitivity loss and trained $C$ to maximize it whereas trained $F$ to minimize it. We also implemented $C$ with deeper layers, but could not find improvement.

\textbf{ENT}. The difference from MME is that the entire network is trained to minimize entropy loss for unlabeled examples in addition to classification loss. 

\textbf{CDAN}~\cite{long2017conditional}. We used the official implementation of CDAN provided in \url{https://github.com/thuml/CDAN}. For brevity, CDAN in our paper denotes CDAN+E in their paper. We changed their implementation so that the model is trained with labeled target examples. Similar to DANN, the domain classifier of CDAN is trained to distinguish source examples and unlabeled target examples.

\section{Additional Results Analysis}
\textbf{Results on Office-Home and Office.}
In Table~\ref{tb:office-home_all} and Table~\ref{tb:office_all}, we report all results on Office-Home and Office. In almost all settings, our method outperformed baseline methods. 

\begin{figure}[htbp]
   \begin{center}
   \includegraphics[width=0.7\hsize]{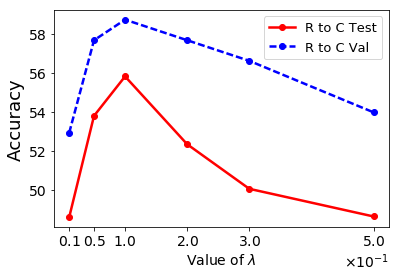}
  \end{center}
  \vspace{-7mm}
   \caption{Sensitivity to hyper-parameter $\lambda$. The result is obtained when we use AlexNet on DomainNet, Real to Clipart.}
   \label{fig:lambda_change}
  
\end{figure}

\textbf{Sensitivity to hyper-parameter $\lambda$.}
In Fig.~\ref{fig:lambda_change}, we show our method's performance when varying the hyper-parameter $\lambda$ which is the trade-off parameter between classification loss on labeled examples and entropy on unlabeled target examples. The best validation result is obtained when $\lambda$ is $0.1$. From the result on validation, we set $\lambda$ 0.1 in all experiments.

\begin{figure}[h!]

\begin{subfigure}{0.49\hsize}
  \begin{center}
   \includegraphics[width=\hsize]{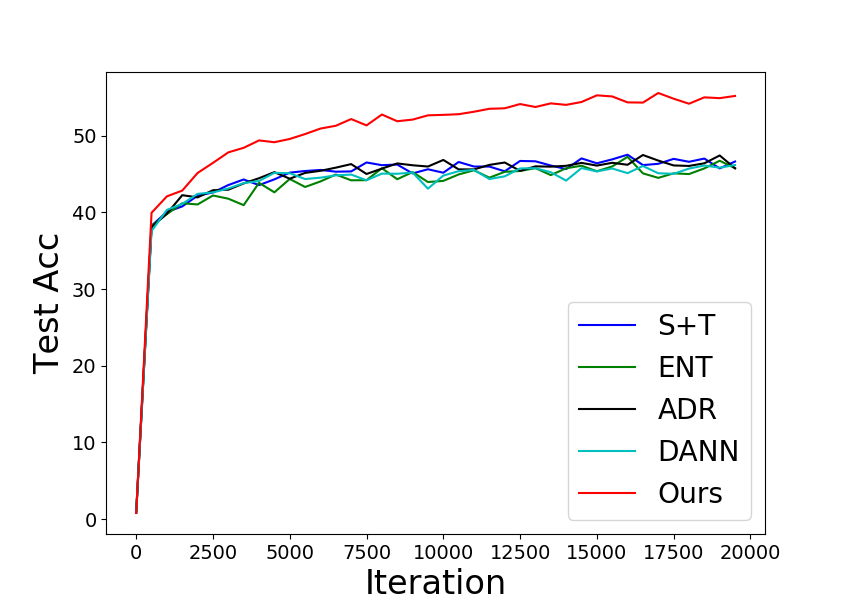}
   \caption{Test accuracy}
   \label{fig:logi}
  \end{center}
  
 \end{subfigure}
 \begin{subfigure}{0.49\hsize}
  \begin{center}
   \includegraphics[width=\hsize]{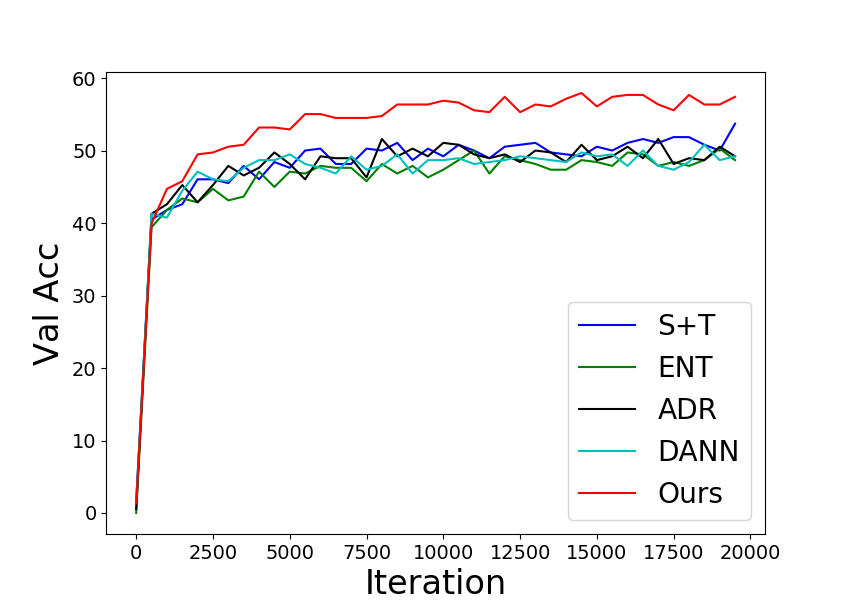}
   \caption{Validation accuracy}
   \label{fig:entropy}
  \end{center}
 \end{subfigure}
  
 \vspace{-3mm}
    \caption{Test and validation accuracy over iterations. Our method increases performances over iterations while others quickly converges. The result is obtained on Real to Clipart adaptation of DomainNet using AlexNet.}
        \label{fig:acc_change}
\end{figure}

\textbf{Changes in accuracy during training.}
We show the learning curve during training in Fig~\ref{fig:acc_change}. Our method gradually increases the performance whereas others quickly converges.  

\begin{table*}[t]
\begin{center}
\scalebox{0.85}{
\begin{tabular}{l|l|cccccccccccc|c}
\toprule[1.5pt] 
Network & Method       &R to C& R to P & R to A & P to R & P to C & P to A & A to P & A to C & A to R & C to R & C to A & C to P & Mean \\\hline
   \multicolumn{15}{c}{\bf{One-shot}} \\\hline
\multirow{6}{*}{AlexNet}&S+T  & 37.5 & 63.1 & 44.8 & 54.3 & 31.7 & 31.5 & 48.8 & 31.1 & 53.3 & 48.5 & 33.9 & 50.8 & 44.1 \\
&DANN & \bf{42.5} & 64.2 & 45.1 & 56.4 & 36.6 & 32.7 & 43.5 & 34.4 & 51.9 & 51.0 & 33.8 & 49.4 & 45.1 \\
&ADR  & 37.8 & 63.5 & 45.4 & 53.5 & 32.5 & 32.2 & 49.5 & 31.8 & 53.4 & 49.7 & 34.2 & 50.4 & 44.5 \\
&CDAN & 36.1 & 62.3 & 42.2 & 52.7 & 28.0 & 27.8 & 48.7 & 28.0 & 51.3 & 41.0 & 26.8 & 49.9 & 41.2 \\
&ENT  & 26.8 & 65.8 & 45.8 & 56.3 & 23.5 & 21.9 & 47.4 & 22.1 & 53.4 & 30.8 & 18.1 & 53.6 & 38.8 \\
&MME  & 42.0 & \bf{69.6} & \bf{48.3} & \bf{58.7} & \bf{37.8} & \bf{34.9} & \bf{52.5} & \bf{36.4} & \bf{57.0} & \bf{54.1} & \bf{39.5} & \bf{59.1} & \bf{49.2} \\\hline\hline
\multirow{6}{*}{VGG}&S+T  & 39.5 & 75.3 & 61.2 & 71.6 & 37.0 & 52.0 & 63.6 & 37.5 & 69.5 & 64.5 & 51.4 & 65.9 & 57.4 \\
&DANN & \bf{52.0} & 75.7 & 62.7 & 72.7 & 45.9 & 51.3 & 64.3 & 44.4 & 68.9 & 64.2 & 52.3 & 65.3 & 60.0 \\
&ADR  & 39.7 & 76.2 & 60.2 & 71.8 & 37.2 & 51.4 & 63.9 & 39.0 & 68.7 & 64.8 & 50.0 & 65.2 & 57.4 \\
&CDAN & 43.3 & 75.7 & 60.9 & 69.6 & 37.4 & 44.5 & 67.7 & 39.8 & 64.8 & 58.7 & 41.6 & 66.2 & 55.8 \\
&ENT  & 23.7 & 77.5 & 64.0 & \bf{74.6} & 21.3 & 44.6 & 66.0 & 22.4 & 70.6 & 62.1 & 25.1 & 67.7 & 51.6 \\
&MME  & 49.1 & \bf{78.7} & \bf{65.1} & 74.4 & \bf{46.2} & \bf{56.0} & \bf{68.6} & \bf{45.8} & \bf{72.2} & \bf{68.0} & \bf{57.5} & \bf{71.3} & \bf{62.7}\\\hline
 \multicolumn{15}{c}{\bf{Three-shot}} \\\hline
 \multirow{6}{*}{AlexNet} & S+T          & 44.6   & 66.7   & 47.7   & 57.8   & 44.4   & 36.1   & 57.6   & 38.8   & 57.0   & 54.3   & 37.5   & 57.9   & 50.0 \\
       
 & DANN         & 47.2   & 66.7   & 46.6   & 58.1   & 44.4   & 36.1   & 57.2   & 39.8   & 56.6   & 54.3   & 38.6   & 57.9   & 50.3 \\
     & ADR          & 45.0   & 66.2   & 46.9   & 57.3   & 38.9   & 36.3   & 57.5   & 40.0   & 57.8   & 53.4   & 37.3   & 57.7   & 49.5 \\
        & CDAN         & 41.8   & 69.9   & 43.2   & 53.6   & 35.8   & 32.0   & 56.3   & 34.5   & 53.5   & 49.3   & 27.9   & 56.2   & 46.2 \\
         & ENT          & 44.9   & 70.4   & 47.1   & 60.3   & 41.2   & 34.6   & 60.7   & 37.8   & 60.5   & 58.0   & 31.8   & 63.4   & 50.9 \\
        & MME & \bf{51.2}   & \bf{73.0}   & \bf{50.3}   & \bf{61.6}   & \bf{47.2}   & \bf{40.7}   & \bf{63.9}   & \bf{43.8}   & \bf{61.4}   & \bf{59.9}   & \bf{44.7}   & \bf{64.7}   & \bf{55.2} \\\hline\hline
\multirow{6}{*}{VGG}     & S+T          & 49.6   & 78.6   & 63.6   & 72.7   & 47.2   & 55.9   & 69.4   & 47.5   & 73.4   & 69.7   & 56.2   & 70.4   & 62.9 \\
       
      & DANN         & 56.1   & 77.9   & 63.7   & 73.6   & 52.4   & 56.3   & 69.5   & 50.0   & 72.3   & 68.7   & 56.4   & 69.8   & 63.9 \\
& ADR          & 49.0   & 78.1   & 62.8   & 73.6   & 47.8   & 55.8   & 69.9   & 49.3   & 73.3   & 69.3   & 56.3   & 71.4   & 63.0 \\
        & CDAN         & 50.2   & 80.9   & 62.1   & 70.8   & 45.1   & 50.3   & 74.7   & 46.0   & 71.4   & 65.9   & 52.9   & 71.2   & 61.8 \\
         & ENT          & 48.3   & 81.6   & 65.5   & 76.6   & 46.8   & 56.9   & 73.0   & 44.8   & \bf{75.3}   & \bf{72.9}   & 59.1   & \bf{77.0}   & 64.8 \\
        & MME & \bf{56.9}   & \bf{82.9}   & \bf{65.7}   & \bf{76.7}   & \bf{53.6}   & \bf{59.2}   & \bf{75.7}   & \bf{54.9}   & \bf{75.3}   & \bf{72.9}   & \bf{61.1}   & 76.3   & \bf{67.6}\\
        \bottomrule[1.5pt]
\end{tabular}}
\end{center}
\vspace{-5mm}
\caption{Results on Office-Home. Our method performs better than baselines in most settings.}
\label{tb:office-home_all}

\end{table*}
\textbf{Comparison with virtual adversarial training.}
Here, we present the comparison with general semi-supervised learning algorithm. We select virtual adversarial training (VAT) ~\cite{miyato2015distributional} as the baseline because the method is one of the state-of-the art algorithms on semi-supervised learning and works well on various settings. The work proposes a loss called virtual adversarial loss. The loss is defined as the robustness of the conditional label distribution around each input data point against local perturbation. We add the virtual adversarial loss for unlabeled target examples in addition to classification loss. We employ hyper-parameters used in the original implementation because we could not see improvement in changing the parameters. 
We show the results in Table~\ref{tb:vat}. We do not observe the effectiveness of VAT in SSDA. This could be due to the fact that the method does not consider the domain-gap between labeled and unlabeled examples. In order to boost the performance, it should be better to account for the gap. 

\textbf{Analysis of Batch Normalization.}
We investigate the effect of BN and analyze the behavior of entropy minimization and our method with ResNet. When training all models, unlabeled target examples and labeled examples are forwarded separately. Thus, the BN stats are calculated separately between unlabeled target and labeled ones. Some previous work~\cite{cariucci2017autodial,li2016revisiting} have demonstrated that this operation can reduce domain-gap. We call this batch strategy as a ``Separate BN''.
To analyze the effect of Separate BN, we compared this with a ``Joint BN'' where we forwarded unlabeled and labeled examples at once. BN stats are calculated jointly and Joint BN will not help to reduce domain-gap. We compare ours with entropy minimization on both Separate BN and Joint BN. Entropy minimization with Joint BN performs much worse than Separate BN as shown in Table \ref{tb:tb_bn}. This results show that entropy minimization does not reduce domain-gap by itself.
On the other hand, our method works well even in case of Joint BN. This is because our training method is designed to reduce domain-gap. 

\textbf{Comparison with SSDA methods~\cite{shu2018dirt, ao2017fast}}
Since there are no recently proposed SSDA methods using deep learning, we  compared with the state-of-the-art unsupervised DA methods modified for the SSDA task. We also compared our method with \cite{shu2018dirt} and \cite{ao2017fast}. We implemented \cite{shu2018dirt} and also modified it for the SSDA task. To compare with~\cite{ao2017fast}, we follow their evaluation protocol and report our and their best accuracy (see Fig. 3 (c)(f) in~\cite{ao2017fast}). As shown in Table~\ref{tb:compare_base}, we outperform these methods with a significant margin. 

\textbf{Results on Multiple Runs.}
We investigate the stability of our method and several baselines.
Table \ref{tb:tb_run} shows results averaged accuracy and standard deviation of three runs. The deviation is not large and we can say that our method is stable.

\textbf{Results on Different Splits.}
We investigate the stability of our method for labeled target examples.
Table \ref{tb:tb_split} shows results on different splits. {\it sp0} correponds to the split we use in the experiment on our paper. For each split, we randomly picked up labeled training examples and validation examples. Our method consistently performs better than other methods. 

\begin{table}[]
\begin{center}
\scalebox{0.85}{
\begin{tabular}{l|l|cccc}
\toprule[1.5pt]
\multirow{2}{*}{Network}      & \multirow{2}{*}{Method}&\multicolumn{2}{c}{W to A} \ & \multicolumn{2}{c}{D to A}  \\
&&1-shot&3-shot&1-shot&3-shot\\\hline
\multirow{6}{*}{AlexNet}& S+T          &50.4& 61.2   & 50.0   &62.4     \\
&DANN         &57.0& 64.4   & 54.5   &65.2\\
&ADR          &50.2& 61.2   & 50.9   &61.4 \\
&CDAN         &50.4&  60.3  &48.5   &61.4\\
&ENT          &50.7& 64.0   & 50.0  &66.2 \\
&MME  &\bf{57.2} &\bf{67.3}   & \bf{55.8}  &\bf{67.8} \\\hline\hline
\multirow{6}{*}{VGG}&S+T          & 69.2   & 73.2 & 68.2   & 73.3 \\
&DANN         & 69.3   & 75.4 & 70.4   & 74.6 \\
&ADR          & 69.7   & 73.3 & 69.2   & 74.1 \\
&CDAN         & 65.9   & 74.4 & 64.4   & 71.4 \\
&ENT          & 69.1   & 75.4 & 72.1   & 75.1 \\
&MME & \bf{73.1}   & \bf{76.3} & \bf{73.6}   & \bf{77.6} \\
 \bottomrule[1.5pt]
\end{tabular}}
\vspace{-3mm}
\caption{Results on Office. Our method outperformed other baselines in all settings.}
\label{tb:office_all}
\end{center}
\end{table}
\begin{table}[h]
\begin{center}
\scalebox{0.7}{
\begin{tabular}{l|ccccccccc}
\toprule[1.5pt] 
 Method       &R to C&R to P&P to C&C to P&C to S&S to P&	R-S	&P to R   \\\hline
S+T  & 47.1&45.0&44.9&35.9&36.4&38.4&33.3&58.7 \\
VAT &46.1&43.8&44.3&35.8&35.6&38.2&31.8&57.7\\
MME  &55.6&49.0&51.7&40.2&39.4&43.0&37.9&60.7  \\
        \bottomrule[1.5pt]
\end{tabular}}
\end{center}
\vspace{-5mm}
\caption{Comparison with VAT~\cite{miyato2015distributional} using AlexNet on DomainNet. VAT does not perform bettern than S+T}
\label{tb:vat}

\end{table}

\begin{table}[h]
\begin{center}
\begin{tabular}{l|c|c}
\toprule[1.5pt] 
 Method       &{Joint BN} &{Separate BN}\\\hline
         ENT          &63.6  &68.9 \\
         MME         &69.5  &69.6 \\
         \bottomrule[1.5pt]
\end{tabular}
\end{center}
\vspace{-5mm}
\caption{Ablation study of batch-normalization. The performance of the ENT method highly depends on the choice of BN while our method shows consistent behavior.}
\label{tb:tb_bn}

\end{table}
\begin{table}[]
\begin{tabular}{l|cccccc}
\toprule[1.5pt] 
\multirow{2}{*}{Method}     &\multicolumn{3}{c}{1-shot} &  \multicolumn{3}{c}{3-shot}      \\
             & sp0  & sp1  & sp2        & sp0  & sp1  & sp2  \\\hline
S+T          & 43.3 & 43.8 & 43.8       & 47.1 & 45.9 & 48.8 \\
DANN         & 43.3 & 44.0 & 45.4       & 46.1 & 43.1 & 45.3 \\
ENT          & 37.0 & 32.9 & 38.2       & 45.5 & 45.4 & 47.8 \\
MME & \bf{48.9} & \bf{51.2} & \bf{51.4}       & \bf{55.6} & \bf{55.0} & \bf{55.8}\\
  \bottomrule[1.5pt]
\end{tabular}
\vspace{-3mm}
\caption{Results on different training splits on DomainNet, Real to Clipart adaptation scenario using AlexNet.}
\label{tb:tb_split}

\end{table}

\begin{table}[t]
\begin{center}
\scalebox{0.85}{
\begin{tabular}{l|cc||c|cc}
\toprule[1.5pt] 
 \multirow{2}{*}{AlexNet}       &\multicolumn{2}{c||}{R to C} &\multirow{2}{*}{AlexNet}&\multicolumn{1}{c}{D to A}&\multicolumn{1}{c}{W to A}\\
         & 1-shot & 3-shot && 1-shot & 1-shot  \\ \hline
         DIRT-T \cite{shu2018dirt}& 45.2 &48.0& GDSDA \cite{ao2017fast} &51.5&48.3 \\
         MME      & \bf{48.9} &\bf{55.6}& MME &\bf{58.5}&\bf{60.4}    \\
           \bottomrule[1.5pt]
\end{tabular}}
\end{center}
\caption{Comparison with ~\cite{shu2018dirt, ao2017fast}.}
\label{tb:compare_base}
\end{table}

\begin{table}[]
\begin{center}

\begin{tabular}{l|cc}
\toprule[1.5pt] 
\multirow{1}{*}{Method}     &\multicolumn{1}{c}{1-shot} &  \multicolumn{1}{c}{3-shot}      \\\hline
CDAN         & 62.9$\pm$1.5 &65.3$\pm$0.1 \\
ENT          & 59.5$\pm$ 1.5&63.6$\pm$1.3 \\
MME &\bf{64.3}$\pm$ 0.8& \bf{66.8}$\pm$0.4\\
  \bottomrule[1.5pt]
\end{tabular}
\vspace{-3mm}
\caption{Results on three runs on DomainNet, Sketch to Painting adaptation scenario using ResNet. }
\label{tb:tb_run}
\end{center}

\end{table}

